\documentclass[11pt,a4paper]{article}
\usepackage{crisis-domain}

\usepackage{multicol}
\usepackage{multirow}

\usepackage{xcolor}

\usepackage{amsfonts}
\usepackage{amsmath}

\usepackage{times}
\usepackage{url}
\usepackage{hyperref}
\usepackage{cleveref}

\usepackage{caption}
\usepackage{subcaption}

\usepackage{latexsym}
\usepackage{graphicx}

\usepackage{booktabs}

\usepackage{flushend}

\date{}

\begin{document}
\maketitle

\begin{abstract}

In recent years, the task of mining important information from social media posts during crises has become a focus of research for the purposes of assisting emergency response (ES). The TREC Incident Streams (IS) track is a research challenge organised for this purpose. The track asks participating systems to both classify a stream of crisis-related tweets into humanitarian aid related information types and estimate their importance  regarding criticality. The former refers to a multi-label information type classification task and the latter refers to a priority estimation task. In this paper, we report on the participation of the University College Dublin School of Computer Science (UCD-CS) in TREC-IS 2021. We explored a variety of approaches, including simple machine learning algorithms, multi-task learning techniques, text augmentation, and ensemble approaches. The official evaluation results indicate that our runs achieve the highest scores in many metrics. To aid reproducibility, our code is publicly available\footnote{\url{https://github.com/wangcongcong123/crisis-mtl}}.

\end{abstract}
\section{Introduction}
\label{sec:intro}

Unexpected Emergencies can cause substantial loss of both life and property if assistance is not available in a timely manner. Recent studies have sought solutions for more efficient emergency response (ES) using computational techniques~\cite{caragea2011classifying,vieweg2012situational,imran2015processing}. Among these works, social media is acknowledged as a promising venue for mining important messages for ES given that some people do tend to seek help by posting messages on social media as a crisis situation unfolds, these messages may contain critical information of relevance to emergency responders~\cite{imran2015processing,McCreadie2019,McCreadie2020}. 

This motivated the Incident streams (IS) track~\cite{McCreadie2019,McCreadie2020}, which challenges the community to explore effective approaches for identifying important messages from user-posted streams on social media during crises. The IS track is a research challenge consisting of two main tasks. The first asks participating systems to classify a stream of crisis-related tweets into humanitarian aid related categories, known as the multi-label information types (ITs) classification task. IS comprises a total of 25 information types that are defined as the categories of possible aid needs in a crisis such as requesting donations, call for search and rescue, reporting weather, etc. The 25 ITs are further divided into two subcategories; 6 are defined as ``actionable'' ITs (e.g., search and rescue) and the remaining 19 are ``non-actionable'' ones (e.g., reporting weather)\footnote{For a full list of the ITs, see the official website at~\url{http://dcs.gla.ac.uk/\~richardm/TREC\_IS/}}. The second task is known as the priority estimation task. It requires participants to estimate the criticality of those tweets that have been classified into ITs. This criticality is represented by a numeric value from 0 to 1 indicating the least to the most importance.   

Having participated in this track since 2019 (the second iteration of the IS track), our system has evolved based on the experience learnt from our prior participations in past TREC-IS editions~\footnote{The IS track normally runs two editions every year and a new test set is annotated and added to the training set after each edition.}. Unlike previous IS editions~\cite{McCreadie2019,McCreadie2020}, TREC-IS 2021 initiated an online leaderboard for participants~\footnote{\url{https://trecis.github.io/}}. It is noted that the leaderboard only reports the performance of participating runs in the 2021A Edition where the test set is partially annotated within events based on pooling by priority (the submitted test tweets are predicted by ITs and sorted by priority score within each event). In the 2021B Edition, the test set comprises the tweets of more annotated events and deeper pooling (new judgements). Hence, the 2021B Edition acted as an enhanced evaluation for the participating runs that had been submitted to the 2021A leaderboard. Given the timeliness of performance feedback from the leaderboard, we explored a wide range of approaches including a Na\"ive Bayes classifier using contextual sentence embeddings as the features, multi-task learning approaches with text augmentations, and an ensemble technique. We found our runs perform consistently well in both A and B editions and in particular our multi-task learning runs and ensemble runs perform the best in many metrics amongst all participating runs. However, the results did not show that text augmentations can bring overall improvements.

\section{Related Work}

Since the launch of TREC-IS, many works have been produced on the topic of crisis tweet classification and priority estimation (CTC-PE). \citet{wangcmu2019} applied Na\"ive Bayes, Support Vector Machine (SVM), Random Forest, and the ensemble of these models with hand-crafted features for CTC-PE. \citet{choi2018cbnu} applied SVM and deep learning models which combine class activation mapping with one-shot learning in convolutional neural networks for CTC-PE. \citet{miyazaki2019label} applied a BiLSTM model for CTC-PE by incorporating the hierarchical structure of labels into the model. \citet{congcong2020cls} applied a BiLSTM model along with pre-trained ELMo embeddings and trainable embeddings as the input features for CTC-PE. \citet{wang2021} fine-tuned BERT~\cite{BERT2018} in a multi-task learning manner for CTC-PE while \citet{wang2021multi} extended the multi-task learning approach to a sequence-to-sequence transformer-based model T5~\cite{raffel2019exploring}. To alleviate the class imbalanced problem, \citet{sharmaimproving2020} applied synonym replacements as well as crisis image labels to augment the original training data. Other techniques such as downsampling the training data or generating new examples via GPT-2 are also found in the literature~\cite{congcong2020cls,hepburn8university}.

\section{Methods and Experiments}
\label{sec:method}
Table~\ref{tab:runs-overview} summarises the runs we submitted to TREC-IS 2021. The major techniques used in the runs are described as follows.

\textbf{ML run}: In this run, we convert each tweet to a representation via pre-trained sentence embeddings (SBERT) models~\cite{reimers-2019-sentence-bert}. Having tested multiple combinations of the available publicly-available pre-trained variants of SBERT~\footnote{\url{https://huggingface.co/sentence-transformers}}, we finally choose \texttt{all-mpnet-base-v2} and \texttt{paraphrase-xlm-r-multilingual-v1} to embed the tweets, where each tweet's representation is the concatenation of outputs of the two models. Similarly, in choosing the downstream classifier, we exhaustively searched over a list of candidates including SVC, logistic regression, decision tree and random forest. We finally used GaussianNB as the downstream classifier as it brought the best result on the development set. Here the classifier is only for IT prediction whereas the priority is simply mapped from the predicted ITs (An IT's mapped priority score is the average priority of all tweets belonging to this IT in the training set). In this approach, priority is assumed to be a function of the IT.

\textbf{Multi-task and ensemble run}. Similar to~\citet{wang2021}, we train a single model for both the downstream IT classification and the priority estimation tasks in a multi-task learning manner. In simple terms, we fine-tuned a pre-trained DeBERTa model~\cite{he2020deberta} jointly on the two tasks through adding a multi-label classification head and a regression head on top of the model. The model is optimised on a linear combination of the cross entropy loss of classification and MSE regression loss. By doing so, the model is capable of making predictions on both tasks for the test tweets with only one input forward at inference time. Based on this idea, we train multiple individual models varying in model size and training data size. Ultimately the individual models consist of a fine-tuned \texttt{deberta-base}, \texttt{deberta-base} with Easy Data Augmentation (EDA)~\cite{wei2019eda} and \texttt{deberta-large}. EDA is used in our system to augment the training data in order to ensure that every IT has at least 500 examples. We apply this augmentation since the original training data is heavily class-imbalanced. Moreover, we adopt the ensemble approach from~\citet{wang2021} to leverage the predictions of individual models for IT classification and priority estimation. The ensemble approach is simple, using the union of predicted ITs by individual models as the final IT prediction and the highest priority among individual priority predictions as the final priority score for test tweets.

\textbf{Ensemble run with post-processing}. Among the pre-defined 25 ITs~\footnote{\url{http://dcs.gla.ac.uk/\~richardm/TREC\_IS/}}, there is an IT called ``Irrelevant''. The multi-label ITs predicted by the above ensemble approach can contain this class along with other ITs. However, a tweet that is classified as ``Irrelevant'' cannot also be labelled with other ITs.  We thus adopt a post-processing step to handle this issue. For any tweet with this type of prediction, we compare the prediction probability for ``Irrelevant'' with the probabilities of other ITs. The tweet is assigned ``Irrelevant'' if its probability for ``Irrelevant'' is greater than all the individual probabilities of the other ITs. Otherwise it is predicted to be one of the other ITs. As a result, the tweet's priority score also becomes $0$ if it is considered to be ``Irrelevant''.

\textbf{Direct-Generation Augmentation (DGA) and Noise Label Annealing (NLA)}. Aside from EDA augmentation, described above, we also explored other augmentation techniques. Inspired by~\citet{wang2021towards} who applied large pre-trained language models to generate training data without any human annotation and model training but though carefully-crafted prompts, we utilise a similar approach using a small number of examples as the prompt. We choose the pre-trained checkpoint \texttt{gpt-neo-2.7B}~\footnote{\url{https://huggingface.co/EleutherAI/gpt-neo-2.7B}} as the generation model and the prompt template is formulated as follows:

\begin{small}
\vspace{0.2cm}
\colorbox{lightgray}{Tweet for help in disaster}\\

\colorbox{lightgray}{Title: \{IT name\}} \\

\colorbox{lightgray}{Content: \{Tweet text\}}

\end{small}
\vspace{0.2cm}
The template constructs a stream of natural language, starting with a task description~\footnote{The task description is carefully chosen based on our preliminary experiments evaluated on the development set.}, followed by the title and content fields, which are replaced by the IT name and the tweet text respectively. This is something we refer to Direct-Generation Augmentation (DGA). In our DGA-based runs, we sampled two examples of non-target ITs from the training data to construct the prompt. To generate a new example for a target IT, we omit the textual part of the content so that the model learns from the prompt (two sampled non-target examples) to complete the content part of the target IT. Finally, we used DGA to augment the training data, thus ensuring that every IT has at least $1000$ examples. One challenge associated with this kind of augmentation is that the generated texts are likely not to be label-aligned with the label it should be and these generated texts are deemed to be noisy or label-incompatible data that is harmful to the downstream task performance. We adopt a strategy called Noisy Label Annealing (NLA) introduced in~\citet{wang2021towards} to filter out noisy training signals as training progresses. The general idea is that we check the predictions of augmented training examples at the end of each epoch of downstream model training and remove an example if the model disagrees with its label with high confidence.

Regarding model training, we remove approximately $10\%$ of the original training data to use as the development set. We fine-tune the multi-task learning model with $10$ epochs and select the best checkpoint based on the IT macro-F1 score on the development set. The model's parameters are tuned on batches (batch size = $16$) of training data using Adam~\cite{kingma2014adam} as the optimizer with a linear warm-up scheduler changing the learning rate from $0$ to $5e-5$ within the first $10\%$ of total training steps and then linearly decays to $0$. Apart from these, the rest of hyper-parameters are set up the same as the default by the transformers library~\cite{Wolf2019HuggingFacesTS}.

\section{Results and Discussions}

\begin{table*}[t]
\small
\centering
\begin{tabular}{ll}
\hline
Run   names & Description                                                                                        \\
\hline
ucdcs-strans.nb      & \textbf{ML run} of using SBERT as the fixed features and GaussianNB as the downstream classifier               \\
ucdcs-run1           & \textbf{Multi-task run} using deberta-base                                      \\
ucdcs-run2           & \textbf{Multi-task run} using deberta-base with \textbf{EDA augmentation}                                           \\
        ucdcs-run3           & \textbf{Multi-task run} using deberta-large                                         \\
ucdcs-mtl.ens (run4) & \textbf{Ensemble run} of run 1, 2 and 3                             \\
ucdcs-mtl.ens.new    & \textbf{Ensemble run} with \textbf{post processing}                                                                     \\
ucdcs-mtl.fta        & \textbf{Multi-task run} of deberta-base with \textbf{direct-generation augmentation (DGA)}                           \\
ucdcs-mtl.fta.nla    & \textbf{Multi-task run} of deberta-base with \textbf{DGA} plus \textbf{noise label annealing (NLA)} \\
ucdcs-mtl.ens.fta    & \textbf{Ensemble run} of run1, 3 and mtl.fta.nla \\
\hline
\end{tabular}

\caption{Overview of UCD-CS runs at TREC-IS 2021. Details of the techniques in bold are elaborated in Section~\ref{sec:method}.}
\label{tab:runs-overview}
\end{table*}

\begin{table*}[]
\centering
\small
\begin{tabular}{lllllllll}
\hline
                     & \textbf{nDCG} & \textbf{IT F1 {[}A{]}} & \textbf{IT F1 {[}All{]}} & \textbf{IT Acc.} & \textbf{Pri F1 {[}A{]}} & \textbf{Pri F1 {[}All{]}} & \textbf{Pri R {[}A{]}} & \textbf{Pri R {[}All{]}} \\
                     \hline
ucdcs-strans.nb      & 0.4297          & 0.2423 & 0.2695          & 0.8294          & 0.1998         & 0.1988          & 0.147  & 0.1514 \\
ucdcs-run1           & \textbf{0.6115} & 0.215  & 0.2951          & 0.8837          & 0.3032         & 0.3068          & 0.2592 & 0.297  \\
ucdcs-run2           & 0.5848          & 0.2215 & 0.2984          & 0.8835          & 0.25           & 0.2781          & 0.2305 & 0.2748 \\
ucdcs-run3           & 0.6051          & 0.2391 & 0.31            & 0.8852          & 0.272          & 0.3066          & 0.3112 & 0.3325 \\
ucdcs-mtl.ens (run4) & 0.5907          & 0.2579 & \textbf{0.3211} & 0.8646          & 0.3052         & 0.3125          & 0.325  & 0.3416 \\
ucdcs-mtl.ens.new    & 0.5951          & 0.2627 & 0.3205          & 0.8686          & 0.305          & \textbf{0.3211} & 0.2892 & 0.3089 \\
ucdcs-mtl.fta        & 0.589           & 0.1986 & 0.2793          & \textbf{0.8902} & 0.2769         & 0.2807          & 0.2471 & 0.3001 \\
ucdcs-mtl.fta.nla    & 0.529           & 0.2007 & 0.2751          & 0.8815          & 0.262          & 0.281           & 0.1721 & 0.2193 \\
ucdcs-mtl.ens.fta    & 0.5755          & 0.1592 & 0.2597          & 0.8034          & \textbf{0.306} & 0.3141          & 0.2786 & 0.2855 \\
\hline
med                  & 0.5695          & 0.206  & 0.2823          & 0.8827          & 0.2113         & 0.2175          & 0.1728 & 0.2099 \\
max                  & \textbf{0.6115} & 0.2815 & \textbf{0.3211} & \textbf{0.8902} & \textbf{0.306} & \textbf{0.3211} & 0.4349 & 0.3585 \\
\hline
\end{tabular}

\caption{The performance of UCD-CS runs at TREC-IS 2021 based on results using only the judgments in 2021A Edition. The figures in \textbf{bold} indicate the best scores across all participating runs. The med and max rows present the median and maximum scores of each metric respectively across all participating runs.}
\label{tab:2021a-results}
\end{table*}

\begin{table*}[]
\centering
\small
\begin{tabular}{lllllllll}
\hline
                     & \textbf{nDCG} & \textbf{IT F1 {[}A{]}} & \textbf{IT F1 {[}All{]}} & \textbf{IT Acc.} & \textbf{Pri F1 {[}A{]}} & \textbf{Pri F1 {[}All{]}} & \textbf{Pri R {[}A{]}} & \textbf{Pri R {[}All{]}} \\
                     \hline
                     ucdcs-strans.nb      & 0.338         & 0.1861                 & 0.2395                   & 0.8557           & 0.1733                  & 0.1688                    & 0.0425                 & 0.1003                   \\
ucdcs-run1           & 0.4499        & 0.2177                 & 0.247                    & 0.8966           & 0.2376                  & 0.2566                    & 0.1547                 & 0.2525                   \\
ucdcs-run2           & 0.4361        & 0.2087                 & 0.2433                   & 0.8947           & 0.242                   & 0.2528                    & 0.207                  & 0.2622                   \\
ucdcs-run3           & 0.4583        & 0.2218                 & 0.2539                   & 0.8964           & 0.2543                  & 0.2756                    & 0.221                  & 0.2571                   \\
ucdcs-mtl.ens (run4) & 0.4521        & 0.2361                 & 0.2591                   & 0.8708           & 0.2753                  & 0.2582                    & \textbf{0.2302}        & \textbf{0.2952}          \\

ucdcs-mtl.ens.new    & 0.4555        & \textbf{0.251}         & \textbf{0.2623}          & 0.8753           & 0.2783                  & 0.2703                    & 0.2116                 & 0.2604                   \\
ucdcs-mtl.fta        & 0.4464        & 0.1958                 & 0.2369                   & \textbf{0.9067}  & 0.2309                  & 0.2333                    & 0.2044                 & 0.2454                   \\
ucdcs-mtl.fta.nla    & 0.4069        & 0.1687                 & 0.2187                   & 0.8945           & 0.2588                  & 0.2635                    & 0.1861                 & 0.2122                   \\
ucdcs-mtl.ens.fta    & 0.4448        & 0.0946                 & 0.1889                   & 0.8113           & \textbf{0.2798}         & 0.2724                    & 0.2149                 & 0.2629                   \\
\hline
med & 0.4272 & 0.1842 & 0.233 & 0.8947 & 0.2107 & 0.2031 & 0.1495 & 0.1993 \\
max         & 0.4791        & \textbf{0.251}         & \textbf{0.2623}          & \textbf{0.9067}  & \textbf{0.2798}         & 0.2756                    & \textbf{0.2302}        & \textbf{0.2952}         \\
\hline
\end{tabular}

\caption{The performance of UCD-CS runs at TREC-IS 2021 based on results using only the judgments in 2021B Edition. The figures in \textbf{bold} indicate the best scores across all participating runs. The med and max rows present the median and maximum scores of each metric respectively across all participating runs.}
\label{tab:2021b-results}
\end{table*}

\begin{table*}[]
\centering
\footnotesize
\begin{tabular}{lllllllll}
\hline
                     & \textbf{nDCG} & \textbf{IT F1 {[}A{]}} & \textbf{IT F1 {[}All{]}} & \textbf{IT Acc.} & \textbf{Pri F1 {[}A{]}} & \textbf{Pri F1 {[}All{]}} & \textbf{Pri R {[}A{]}} & \textbf{Pri R {[}All{]}} \\
                     \hline
                     ucdcs-strans.nb      & 0.3368            & 0.2083                                 & 0.2575                          & 0.8474                      & 0.1959                                & 0.1712                         & 0.1096                               & 0.1417                        \\
ucdcs-run1           & 0.4727            & 0.2433                                 & 0.2772                          & 0.8926                      & 0.2657                                & 0.2632                         & 0.259                                & 0.2888                        \\
ucdcs-run2           & 0.4569            & 0.2326                                 & 0.2753                          & 0.8911                      & 0.2536                                & 0.2524                         & 0.1995                               & 0.2686                        \\
ucdcs-run3           & 0.4707            & 0.2538                                 & 0.286                           & 0.893                       & 0.253                                 & 0.2694                         & 0.2741                               & 0.3053                        \\
ucdcs-mtl.ens (run4) & 0.4617            & 0.267                                  & 0.2923                          & 0.8685                      & 0.2817                                & 0.2623                         & 0.2886                               & \textbf{0.3182}               \\

ucdcs-mtl.ens.new    & 0.4643            & \textbf{0.2784}                        & \textbf{0.2946}                 & 0.8728                      & \textbf{0.2864}                       & \textbf{0.2734}                & 0.2748                               & 0.2827                        \\
ucdcs-mtl.fta        & 0.463             & 0.2159                                 & 0.2647                          & \textbf{0.9016}             & 0.2497                                & 0.2361                         & 0.2779                               & 0.2834                        \\
ucdcs-mtl.fta.nla    & 0.4193            & 0.1936                                 & 0.2488                          & 0.8907                      & 0.2727                                & 0.2609                         & 0.265                                & 0.2339                        \\
ucdcs-mtl.ens.fta    & 0.4515            & 0.1131                                 & 0.217                           & 0.8073                      & 0.2852                                & 0.2724                         & 0.2479                               & 0.2665                        \\

\hline
med & 0.4381 & 0.2008 & 0.26 & 0.8911 & 0.2086 & 0.2044 & 0.2087 & 0.2431 \\

max         & 0.4904            & \textbf{0.2784}                        & \textbf{0.2946}                 & \textbf{0.9016}             & \textbf{0.2864}                       & \textbf{0.2734}                & 0.3072                               & \textbf{0.3182}   \\
\hline
\end{tabular}
\caption{The performance of UCD-CS runs at TREC-IS 2021 based on results using the judgments in 2021A and 2021B Editions. The figures in \textbf{bold} indicate the best scores across all participating runs. The med and max rows present the median and maximum scores of each metric respectively across all participating runs.}
\label{tab:all-results}
\end{table*}

In order to measure a system's performance from different perspectives, the IS track defines multiple metrics. The metrics can be broadly divided into two categories: IT classification measurements and priority estimation measurements. They are described as follows:

\begin{itemize}
    \item \textbf{IT classification measurements}: To measure the performance of ITs classification, three metrics are defined. They are IT F1 [A], IT F1 [All] and IT Acc., referring to the F1 score of only actionable ITs classification, the F1 score and the accuracy of all 25 ITs classification respectively.
    
    \item \textbf{Priority estimation measurements}: There are five metrics related to the evaluation of priority estimation. Four of them are: Pri F1 [A], Pri F1 [All], Pri R [A] and Pri R [All], referring to the F1 scores and recall scores of only actionable and all ITs classification respectively. Besides these, nDCG is a ranking metric included in this category to measure a run's average performance in ranking the top 100 test tweets per event by priority.
\end{itemize}

As TREC-IS 2021 has been run with two editions (A and B) that produce two sets of judgements, we report our runs' performance separately in the judgements of each edition as well as in the combined judgements of both editions, as presented in Tables~\ref{tab:2021a-results},~\ref{tab:2021b-results} and~\ref{tab:all-results}.

First in overview, most of our runs perform well consistently in both editions across the participating runs. When compared to the median and maximum of each metric, we find that our multi-task and ensemble runs in particular achieve strong performance, hitting the best scores in many cases. To examine the figures by task, we notice that some of our runs can perform well in one task while under-performing in the other. For example, the \texttt{ML run} achieves decent scores in IT classification but its scores for priority estimation are relatively poor. This is likely due to the simple mapping from IT predictions to priority estimation in that run. We expect the \texttt{ML run} to be improved upon by modelling not just the IT classification but also the priority estimation (a regression task).

In terms of our multi-task runs, we find that these runs tend to achieve strong scores in both tasks. This indicates that the joint learning on both tasks through fine-tuning pre-trained language models (DeBERTa in our case) can help achieve strong performance, which adds support to the results in~\cite{wang2021} in previous TREC-IS editions. Also unsurprisingly, the bigger fine-tuned model brings slightly-improved performance when comparing our \texttt{run3} to our \texttt{run1}. Regarding our ensemble runs without augmentation, they outperform our other runs in almost every metric for both tasks as well as achieving highest scores in many metrics among the participating runs. It is noted the \texttt{mtl.ens.new} run with post-processing (to deal with the ``Irrelevant'' IT) further improves the performance in IT classification as compared to \texttt{run4}. 

To check the effect of EDA augmentation when comparing \texttt{run2} to \texttt{run1}, we see marginal improvements in priority estimation in 2021B (Table~\ref{tab:2021b-results}) but not in the other two tables. Hence, it is difficult to conclude whether the EDA augmentation adds benefit in this scenario. we also see similar results for our DGA-based runs (see our \texttt{mtl.fta}, \texttt{mtl.fta.nla} and \texttt{mtl.ens.fta} runs). This seems somewhat inconsistent with the previous study~\cite{congcong2020cls}. We attribute these results to two possible reasons. First, the training data has grown from around 5,000 to 50,000 since then, so it is quite possible that the advantage of text augmentation in a low-data situation is more obvious. Second, 
since the downstream model used in our current runs is pre-trained on big general text data, the new examples generated by text augmentation may be noisy as well as be redundant (the model learns general language features at pre-training and is likely to augment similar examples itself implicitly during fine-tuning). Hence, better approaches for de-noising and diversifying the augmented examples are avenues of research that we seek to explore in the future.

\section{Conclusion}

In this paper, we report UCD-CS's participation at the TREC 2021 Incident Streams track (TREC-IS). We submitted multiple runs  and our approaches included machine learning algorithms, multi-task learning techniques and ensemble approaches. Among these runs, we find in particular that our multi-task and ensemble runs achieve strong performance in both the information type classification and priority estimation tasks through two rounds of evaluation: TREC-IS 2021A and B editions. Although we explored some text augmentation approaches with the intent of boosting the performance, the results did not indicate consistent performance improvements and thus we seek better augmentation techniques in the future.
\bibliography{crisis-domain}
\bibliographystyle{acl_natbib} 
\appendix
\end{document}